%% file: compositional_generative_model.tex
\documentclass{article}

\usepackage{microtype}
\usepackage{graphicx}
\usepackage{subfigure}
\usepackage{booktabs} %

\usepackage{hyperref}

\input{packages}

\input{macros}

\input{math_commands}

\usepackage[accepted]{icml2024}

\icmltitlerunning{Compositional Generative Modeling: A Single Model is Not All You Need}

\begin{document}

\twocolumn[
\icmltitle{Compositional Generative Modeling: A Single Model is Not All You Need}

\icmlsetsymbol{equal}{*}

\begin{icmlauthorlist}
\icmlauthor{Yilun Du}{mit}
\icmlauthor{Leslie Kaelbling}{mit}

\end{icmlauthorlist}

\icmlaffiliation{mit}{MIT}

\icmlcorrespondingauthor{Yilun Du}{yilundu@mit.edu}

\icmlkeywords{Generative Modeling, Modularity, Compositionality, Energy Based Models}

\vskip 0.3in
]

\printAffiliationsAndNotice{} 

\setlength{\abovedisplayskip}{4pt}
\setlength{\belowdisplayskip}{4pt}

\input{sections/0_abstract}
\input{sections/1_introduction}

\input{sections/2_computational_efficency}

\input{sections/3_new_distributions}

\input{sections/4_discovering_structure}

\input{sections/5_implementation}
\input{sections/6_discussion}

\input{sections/7_acknowledgements}

\bibliography{references}
\bibliographystyle{icml2024}

\input{sections/appendix}

\end{document}

%% file: packages.tex
\usepackage{booktabs}
\usepackage{colortbl}

\usepackage{graphicx}
\usepackage{float,wrapfig}

\usepackage{array}
\usepackage{adjustbox}
\usepackage{multirow}
\usepackage{colortbl}

\usepackage{caption}
\usepackage{color,xcolor}
\usepackage{epsfig}
\usepackage{bbm}
\usepackage{hhline}

\usepackage{amsmath,amsfonts,amsthm,amssymb}
\usepackage{mathtools}
\usepackage{bm}
\usepackage{nicefrac}
\usepackage{microtype}

\usepackage{changepage}
\usepackage{extramarks}
\usepackage{fancyhdr}
\usepackage{lastpage}
\usepackage{setspace}
\usepackage{soul}
\usepackage{xspace}

\usepackage{url}
\usepackage{algorithm, algorithmic}
\usepackage{todonotes} %

\usepackage{titlesec}

%% file: macros.tex
\newcommand{\sect}[1]{Section~\ref{#1}}

\newcommand{\eqn}[1]{Equation~\ref{#1}}
\newcommand{\fig}[1]{Figure~\ref{#1}}

\newcommand{\myparagraph}[1]{{\bf #1}\quad}

\DeclarePairedDelimiterX{\infdivx}[2]{(}{)}{%
  #1\;\delimsize|\delimsize|\;#2%
}

\definecolor{MyDarkBlue}{rgb}{0,0.08,1}
\definecolor{MyDarkGreen}{rgb}{0.02,0.6,0.02}
\definecolor{MyDarkRed}{rgb}{0.8,0.02,0.02}
\definecolor{MyDarkOrange}{rgb}{0.40,0.2,0.02}
\definecolor{MyPurple}{RGB}{111,0,255}
\definecolor{MyRed}{rgb}{1.0,0.0,0.0}
\definecolor{MyGold}{rgb}{0.75,0.6,0.12}
\definecolor{MyDarkgray}{rgb}{0.66, 0.66, 0.66}

%% file: math_commands.tex
\usepackage{amsmath,amsfonts,bm}

\def\eqref#1{equation~\ref{#1}}

\def\1{\bm{1}}

\DeclareMathAlphabet{\mathsfit}{\encodingdefault}{\sfdefault}{m}{sl}
\SetMathAlphabet{\mathsfit}{bold}{\encodingdefault}{\sfdefault}{bx}{n}

\def\gC{{\mathcal{C}}}

\def\gG{{\mathcal{G}}}

\def\gU{{\mathcal{U}}}
\def\gV{{\mathcal{V}}}

%% file: sections/0_abstract.tex
\begin{abstract}
Large monolithic generative models trained on massive amounts of data have become an increasingly dominant approach in AI research. We argue that we should instead construct large generative systems by composing smaller generative models together. We show how such a compositional generative approach enables us to learn distributions in a more data-efficient manner, enabling generalization to parts of the data distribution unseen at training time. We further show how this enables us to program and construct new generative models for tasks completely unseen at training. Finally, we show that in many cases, we can discover compositional components from data. 
\end{abstract}

%% file: sections/1_introduction.tex
\section{Introduction}

In the past two years, increasingly large generative models have become a dominant force in AI research, with compelling results in natural language~\citep{brown2020gpt3}, computer vision~\citep{rombach2022highresolution} and decision-making~\citep{reed2022generalist}. Much of the AI research field has now focused on scaling and constructing increasingly large generative models~\citep{hoffmann2022training}, developing tools to build even larger models~\citep{dao2022flashattention, kwon2023efficient}, and studying how properties emerge as these models scale in size~\citep{lu2023emergent, schaeffer2023emergent}.

Despite significant scaling in generative models, existing models remain far from intelligent, exhibiting poor reasoning ability~\citep{tamkin2021understanding}, extensive hallucinations~\citep{zhang2023siren}, and poor understanding of commonsense relationships in images (\fig{fig:large_model_issue})~\citep{OpenEQA2023}. Despite this, large models have already been trained on most of the existing data on the Internet and have reached the limits of modern computational hardware, costing hundreds of millions of dollars to train (\fig{fig:teaser}). Inference costs of such gigantic models are also prohibitive, requiring large computational clusters and a cost of several dollars for longer queries and answers~\citep{OpenAI}.

In addition, adapting such large models to new task distributions is difficult. Directly fine-tuning larger models is often prohibitively expensive, requiring a large computation cluster and an often difficult-to-acquire fine-tuning dataset. Other works have explored leveraging language and a set of in-context examples to teach models new distributions, but such adaptation is limited to settings that are well expressed using a set of language instructions that are further roughly similar to the distributions already seen during training ~\citep{yadlowsky2023pretraining}.

\looseness=-1
In this paper, we argue that as an alternative to studying how to scale and construct increasingly large monolithic generative models, {\bf we should instead construct complex generative models {\it \bf compositionally} from simpler models}. Each constituent model captures the probability distribution of a subset of variables of the distribution of interest, which are combined to model the more complex full distribution. Individual distributions are therefore much simpler and computationally modeled with both fewer parameters and learnable from less data. Furthermore, the combined model can generalize to unseen portions of the data distribution as long as each constituent dimension is locally in distribution.

\input{figText/teaser}

Such compositional generative modeling enables us to effectively represent the sparsity and symmetry naturally found in nature. Sparsity of interactions, for instance between an agent and external environment dynamics can be encoded by representing each with separate generative models.  Sources of symmetry can be captured using multiple instances of the same independent generative component to represent each occurrence of the symmetry, for instance by tiling patch-level generative model over the patches in an image. Compositional structure is widely used in existing work, to tractably represent high dimensional distributions in Probablistic Graphical Models (PGMs) ~\citep{koller2009probabilistic}, and even in existing generative models, {\it i.e.} autoregressive models which factorize distributions into a set of conditional probability distributions (represented by a single model).

Compositional generative modeling further enables us to effectively program and construct new generative systems for unseen task distributions. Individual generative models can be composed in new ways, with each model specifying a set of constraints, and probabilistic composition seen as a communication language among models, ensuring a distribution is a constructed so that all constraints are satisfied to form the task distribution of interest.  
Such programming further requires no {\it explicit training} or {\it data}, enabling generalization in inference even on distributions with no previously seen data. We illustrate how such recombination enables generalization to new task distributions in decision making, image and video synthesis.

\input{figText/large_model_issue}

The underlying compositional components in generative modeling can in many cases be directly inferred and discovered in an unsupervised manner from data, representing compositional structure such as objects and relations. Such discovered components can then be similarly recombined to form new distributions -- for instance, objects components discovered by one generative model on one dataset can be combined with components discovered by a separate generative model on another dataset to form hybrid scenes with objects in both datasets. We illustrate the efficacy of such discovered compositional structure across domains in images and trajectory dynamics.

Overall, in this paper, we advocate for the idea that we should construct complex generative systems by representing them as a compositional system of simpler components and illustrate its benefits across various domains.

%% file: figText/teaser.tex
\begin{figure}[t]
    \centering
    \includegraphics[width=0.9\linewidth]{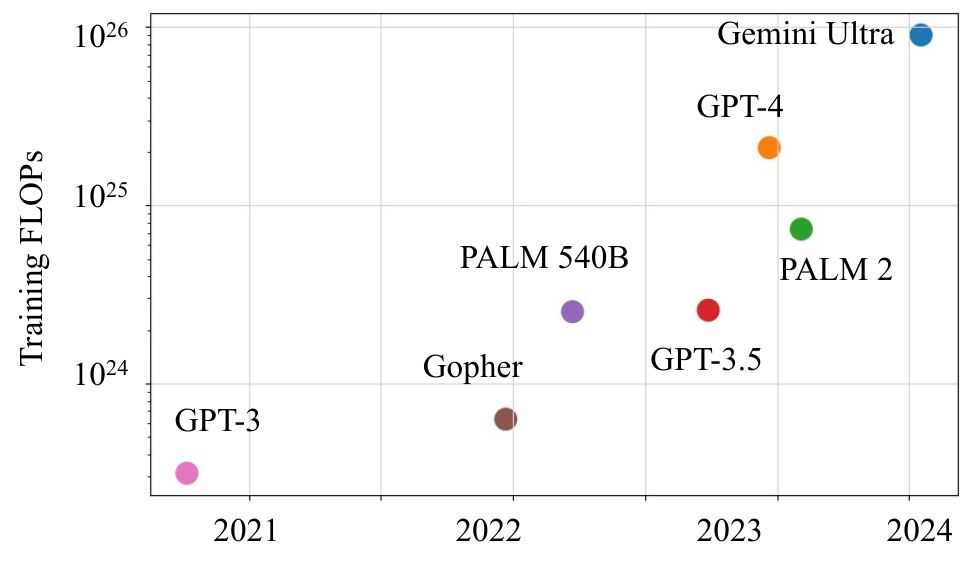}
    \vspace{-10pt}
    \caption{\textbf{Rising Size and Cost of Models.}  While much of AI research has focused on constructing increasingly larger monolithic models, training costs are exponentially rising by a factor of 3 every year with current models already costing several hundred million dollars per training run. Data from \citep{epoch2023aitrends}.}
    \label{fig:teaser}
\vspace{-20pt}
\end{figure}

%% file: figText/large_model_issue.tex
\begin{figure}[t]
    \centering
    \includegraphics[width=\linewidth]{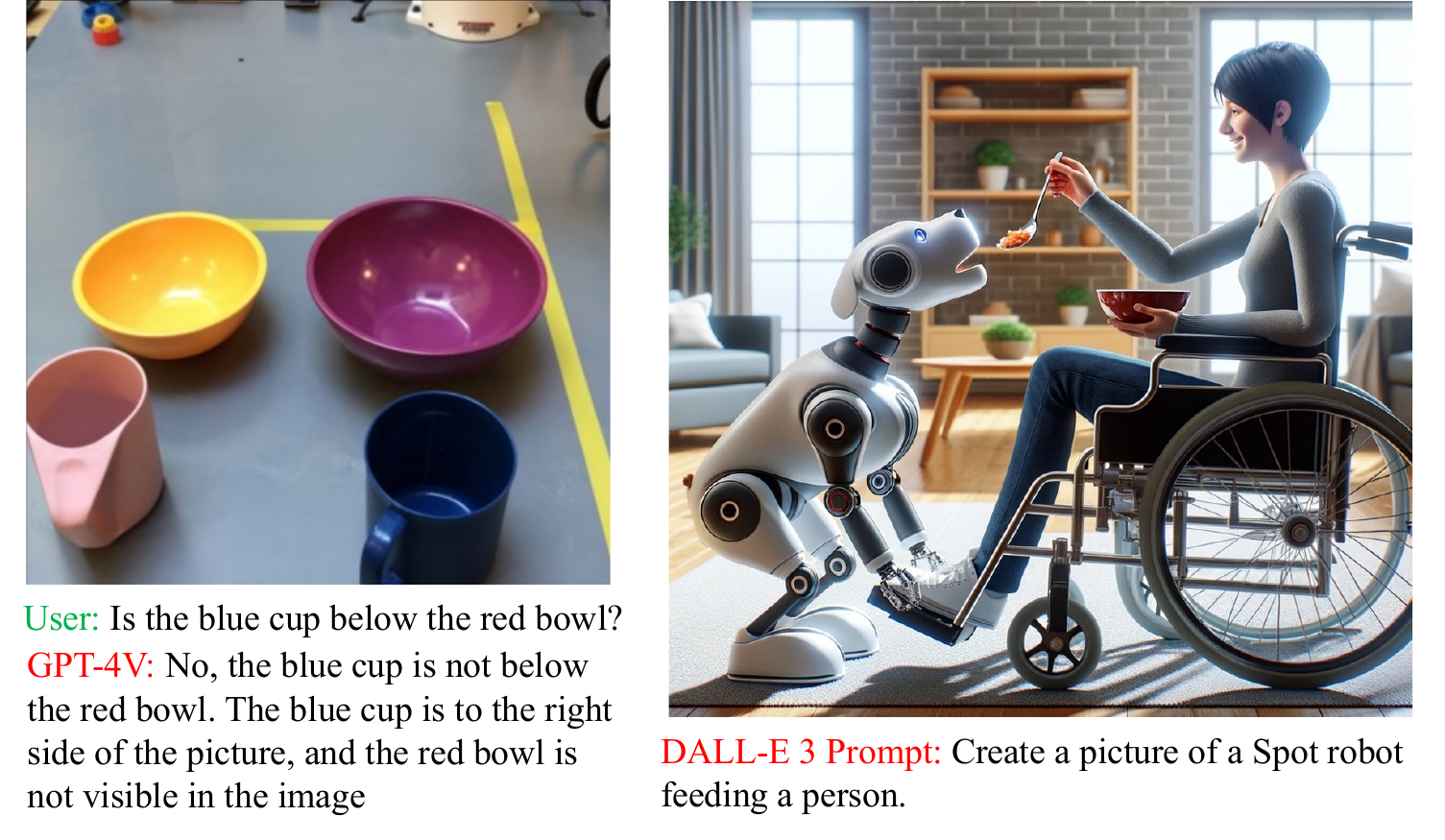}
    \vspace{-25pt}
    \caption{\textbf{Limited Compositionality in Multimodal Models.}  Existing large multimodal models such as GPT-4V and DALL-E 3 still struggle with simple textual queries, often falling back to biases in data.}
    \label{fig:large_model_issue}
\vspace{-15pt}
\end{figure}

%% file: sections/2_computational_efficency.tex
\section{Data Efficient Generative Modeling}

The predominant paradigm for training generative models has been to construct
increasingly larger monolithic models trained with greater
amounts of data and computational power. While language models have
demonstrated significant improvements with increased scale (albeit still with difficulty in
compositionality~\citep{dziri2023faith}), current multimodal models such
as DALL-E 3 and GPT-4V remain unable to take advantage of even simple forms of compositionality (Figure~\ref{fig:large_model_issue}). Such models may be unable to accurately
generate images given combinations of relations rarely seen in training data, or
fail to understand simple spatial relations in images, despite being trained on a
very significant portion of the existing Internet.

One difficulty is that the underlying sample complexity of learning
generative models over joint distributions of variables increases
dramatically with the number of variables. As an example, consider learning probability distributions by maximizing log-likelihood over a set
of random variables A, B, C, D, each of which can take a set of $K$
values. Directly learning a distribution over a single variable A by requires  $O(K)$ values~\cite{canonne2020short}. The data required to learn distributions over a joint set of variables generally increases
exponentially -- so that learning a joint distribution $p(A, B, C, D)$ requires $O(K^4)$
samples~\cite{canonne2020short}.

Constructing large multimodal generative models such as GPT-4V or DALL-E 3 falls into the same
difficulty -- as the number of modalities jointly modeled increases, the
combination of samples required to see and learn the entire data distribution exponentially
increases. This is particularly challenging in the multimodal setting as the existing
data on the Internet used to train these models is often highly non-uniform,
with many combinations of natural language and images unseen.

One approach to significantly reduce the data necessary to
learn generative models over complex joint distributions is factorization --
if we know that a distribution exhibits an independence structure between
variables such as 
\begin{equation*}
    p(A, B, C, D) \propto p(A) p(B) p(C, D),
\end{equation*}
we can substantially reduce the data requirements by only needing to learn these factors, composing them together to form a more complex distribution. This also enables our learned joint
distribution to generalize to unseen combinations of variables so long as
each local variable combination is in distribution (illustrated in \fig{fig:generalize_data}). Even in settings where
 distributions are not accurately modeled as a product  of
 independent factors, such a factorization can still lead to a better models given limited data by reducing the hypothesis space~\cite{murphy2022probabilistic}. This idea of factorizing probability distributions has led to substantial work in probabilistic graphical models (PGMs)~\cite{koller2009probabilistic}. 

Below, we illustrate across four settings how representing a target distribution $p(x)$ in a factorized manner can substantially improve generative modeling performance from a limited amount of data:

\input{figText/generalize_data}

\input{figText/2d_example}

\input{figText/traj_factor_mono}

\myparagraph{Simple Distribution Composition.} In \fig{fig:2d_composition}, we consider modeling a distribution $p(x)$ that is a product $p(x) \propto p_1(x)p_2(x)$ or mixture $p(x) \propto p_1(x) + p_2(x)$ of two factors $p_1(x)$ and $p_2(x)$. We compare training either a single model on $p(x)$ or learning two generative models on the factors $p_1(x)$ and $p_2(x)$. We find that training compositional models leads to a more accurate distribution modeling if the same amount of data is used to learn $p(x)$ as is used to learn both $p_1(x)$ and $p_2(x)$. Even when modeling simple distributions, the data complexity of modeling each factor is simpler than representing the joint distribution.

\myparagraph{Trajectory Modeling.} Next, we consider modeling a probability distribution $p(\tau)$ over trajectories $\tau = (s_0, a_0, s_1, a_1, \ldots, s_T, a_T)$, which many recent works have typically modeled using a single joint distribution $p(s_0, a_0, \ldots, s_T, a_T)$~\citep{janner2022planning, ajay2022conditional}. In contrast to a monolithic generative distribution, given structural knowledge of the environment -- i.e., that it is a Markov Decision Process, a more factorized generative model to represent the distribution is as a product
\begin{equation*}
    p(\tau) \propto \prod_i p(s_i \mid s_{i-1}, a).
\end{equation*}
In \fig{fig:traj_factor_mono}
, we explore the efficacy of compositional and monolithic models in characterizing trajectories in Maze2D, which consists of a 4D state space (2D position and velocity) and 2D action space (2D forces),  using the model in ~\citep{janner2022planning} (with the compositional model representing trajectory chunksize 8 to ensure compatibility with the architecture). We plot the accuracy of generated trajectories at unseen start states as the function of the number of agent episodes used to train models, where each episode has length of approximately 10000 timesteps. As seen in the \fig{fig:traj_factor_mono}(a), given only a very limited number of agent episodes in an environment, a factorized model can more accurately simulate trajectory dynamics.  In addition, we found that training a single joint generative model also took a substantially larger number of iterations to train than the factorized model as illustrated in \fig{fig:traj_factor_mono}(b). 

\input{figText/compositional_visual}

\myparagraph{Compositional Visual Generation.} We further consider modeling a probability distribution $p(x \mid T)$ in text-to-image synthesis, where $x$ is an image and $T$ is a complex text description. While this distribution is usually characterized by a single generative model, we can factor the generation as a product of distributions~\citep{liu2022compositional} given sentences $t_1$, $t_2$, and $t_3$ in the description $T$
\begin{equation*}
    p(x \mid T) \propto p(x \mid t_1) p(x \mid t_2) p(x \mid t_3).
\end{equation*}
This representation of the distribution is more data efficient: we only need to see the full distribution of images given single sentences. In addition, it enables us to generalize to unseen regions of $p(x \mid T)$ such as unseen combinations of sentences and longer text descriptions. In \fig{fig:composition_visual}, we illustrate the efficacy of such an approach.

\myparagraph{Composing Language Models.} Finally, we consider modeling a probability distribution $p(x)$ over a language sequence $x$. Similar to the previous examples, we can represent the likelihood as a composition $p(x) \propto \prod_i p_i(x)$, where each distribution $p_i(x)$ is parameterized by a separate language model. However, directly sampling from such a composition of language models is difficult as it requires intermediate access to the output logits of each model, which are often unavailable for proprietary models. One approach to avoid this issue is to combine outputs of individual language models $p_i(x)$ in the language space and use the result as context for representing the final distribution $p(x)$ over language sequences~\citep{du2023improving}.

In ~\citet{du2023improving}, this compositional approach is found to effectively improve the performance of base language models. For instance, on the MATH dataset~\citep{hendrycks2021measuring}, by composing 5 instances of a GPT-3.5 model, we can obtain a final accuracy of $58.0 \pm 2.8\%$, even outperforming a much larger and expensive GPT-4 model, which obtains a performance of $55.0 \pm 2.9\%$.

%% file: figText/generalize_data.tex
\begin{figure}[t]
\includegraphics[width=\linewidth]{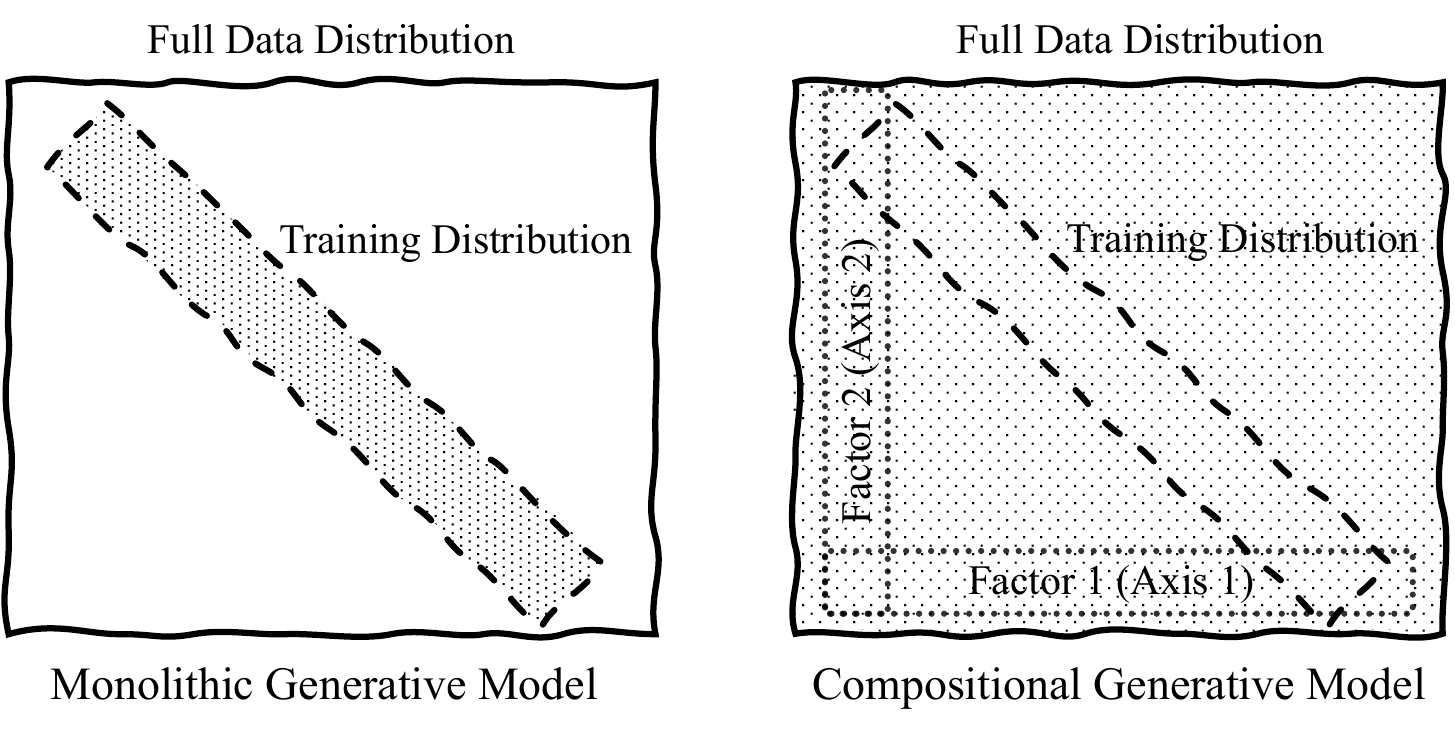}
\vspace{-25pt}
\caption{\small \textbf{Generalizing Outside Training Data.} Given a narrow slice of training data, we can learn generative models that generalize outside the data through composition. We learn separate generative models to model each axis of the data -- the composition of models can then cover the entire data space.}
\label{fig:generalize_data}
\vspace{-10pt}
\end{figure}

%% file: figText/2d_example.tex
\begin{figure}[t]
\includegraphics[width=\linewidth]{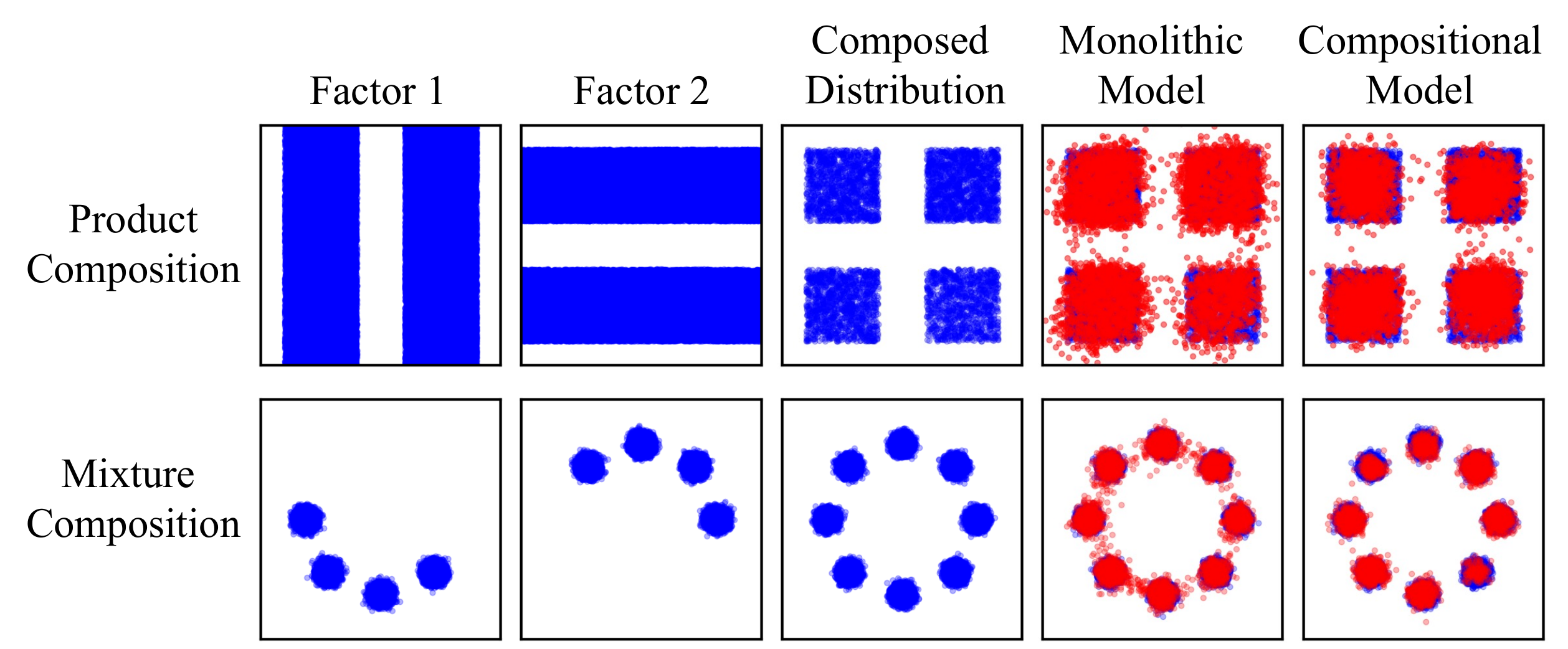}
\vspace{-20pt}
\caption{\small \textbf{Distribution Composition} -- When modeling simple product (top) or mixture  (bottom) compositions, learning two compositional models on the factors is more data efficient than learning a single monolithic model on the product distribution. The monolithic model is trained on twice as much data as individual factors.}
\label{fig:2d_composition}
\vspace{-10pt}
\end{figure}

%% file: figText/traj_factor_mono.tex
\begin{figure}[t]
\includegraphics[width=\linewidth]{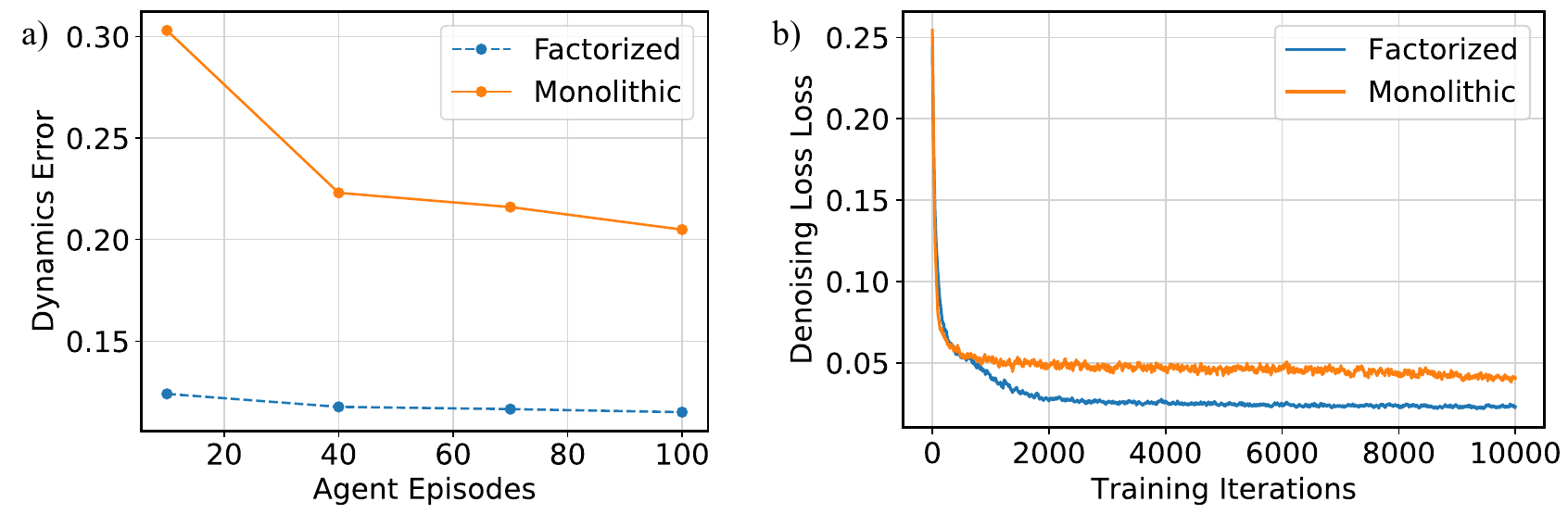}
\vspace{-20pt}
\caption{\small \textbf{Compositional Trajectory Generation} -- By factorizing a trajectory generative model into a set of components, models are able to more accurately simulate dynamics from limited trajectories (a) and train in fewer training iterations (b).}
\label{fig:traj_factor_mono}
\vspace{-5pt}
\end{figure}

%% file: figText/compositional_visual.tex
\begin{figure}[t]
    \centering
    \includegraphics[width=\linewidth]{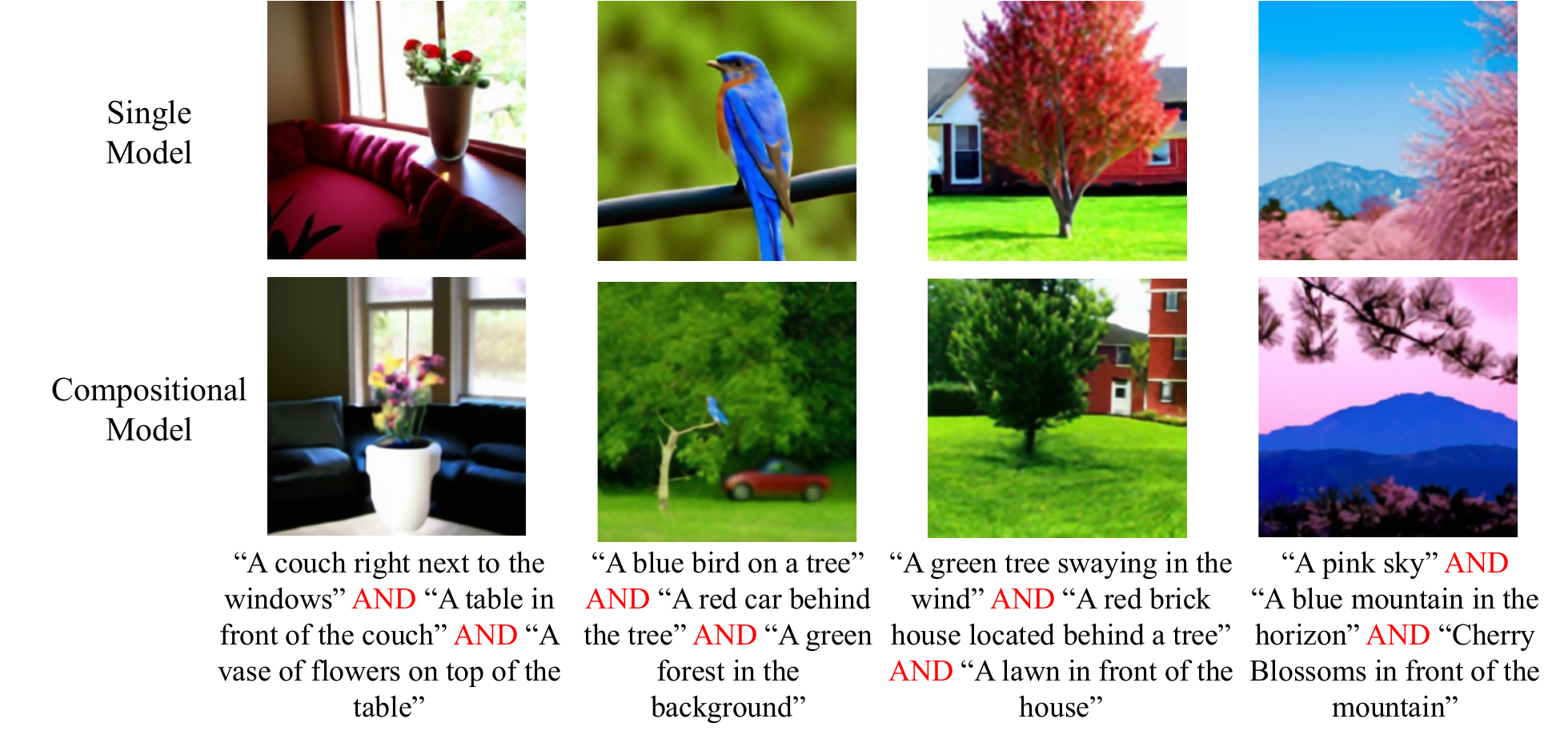}
    \vspace{-20pt}
    \caption{\textbf{Compositional Visual Synthesis.}  By composing a set of generative models modeling conditional image distributions given a sentence description, we can more accurately synthesize images given paragraph-level text descriptions. Figure adapted from ~\citep{liu2022compositional}}
    \label{fig:composition_visual}
\vspace{-15pt}
\end{figure}

%% file: sections/3_new_distributions.tex
\section{Generalization to New Distributions}

In the previous section, we've illustrated how composition can enable us to effectively model a distribution $p(x)$, including areas we have not seen any data in. In this section, we further illustrate how composition enables generalization, allowing us to re-purpose a generative model $p(x)$ to solve a new task by constructing a new generative model $q(x)$.

\input{figText/planning_composition}

Consider the task of planning, where we wish to construct a generative model $q(\tau)$ which samples plans that reach a goal state $g$ starting from a start state $s$. Given a generative model $p(\tau)$, which sample legal, but otherwise unconstrained, state sequences in an environment, we can construct an additional generative model $r(\tau, s, g)$ which has high likelihood when $\tau$ has start state $s$ and goal state $g$ and low likelihood everywhere else. By composing the two distributions
\begin{equation}
    \label{eqn:plan_compose}
    q(\tau) \propto p(\tau) r(\tau, s, g),
\end{equation}
we can construct our desired planning distribution $q(\tau)$, exploiting the fact that probability can be treated as a ``currency" to combine models, enabling us to selectively choose trajectories that satisfy the constraints in both distributions.

Below, we illustrate a set of  applications where we can construct new compositional generative models $q(x)$   to solve tasks in planning, constraint satisfaction, hierarchical decision-making, and image and video generation.

\myparagraph{Planning with Trajectory Composition.} We first consider constructing $q(\tau)$ representing planning as described in \eqn{eqn:plan_compose}. In \fig{fig:compose_plan} we illustrate how sampling from this composed distribution enables successful planning from start to goal states. Quantatively, this approach performs well also as illustrated in~\citep{janner2022planning}. 

\input{figText/csp_example.tex}
\myparagraph{Manipulation through Constraint Satisfaction.} We next illustrate how we can construct a generative model $q(V)$ to solve a variety of robotic object arrangement tasks. As illustrated in \fig{fig:csp-example}, many object arrangement tasks can be formulated as {\it continuous constraint satisfaction problems} consisting of a graph $\gG = \langle \gV, \gU, \gC \rangle$, where $v \in \gV$ is a decision variable (such as the pose of an object), while each $u \in \gU$ is a conditioning variable (such as the geometry of an object) and $c \in \gC$ is a constraint such as collision-free.  Given such a specification, we can solve the robotics tasks by sampling from the composed distribution
\begin{equation*}
    q(V) \propto \prod_{c \in \gC} p_c(\gV^{c} \mid \gU^{c}),
\end{equation*}
corresponding to solving the constraint satisfaction problem. Such an approach enables effective generalization to new problems~\cite{yang2023compositional}, to temporally extended plans~\cite{mishra2023generative}, and the combination of heterogenous policies~\cite{wang2024poco}.

\input{figText/hier_composition}
\input{figText/image_tapestry}

\myparagraph{Hierarchical Planning with Foundation Models.} We further illustrate how we can construct a generative model that functions as a hierarchical planner for long-horizon tasks.  We construct $q(\tau_{\text{text}}, \tau_{\text{image}}, \tau_{\text{action}})$, which jointly models the distribution over a text plan $\tau_{\text{text}}$, image plan $\tau_{\text{image}}$, and action plan $\tau_{\text{action}}$ given a natural language goal $g$ and image observation $o$, by combining pre-existing foundation models trained on Internet knowledge. We formulate $q(\tau_{\text{text}}, \tau_{\text{image}}, \tau_{\text{action}})$ through the composition
\begin{equation*}
   p_{\text{LLM}}(\tau_{\text{text}}, g)  p_{\text{Video}}(\tau_{\text{image}}, \tau_{\text{text}}, o) p_{\text{Action}}(\tau_{\text{action}}, \tau_{\text{image}}).
\end{equation*}
This distribution assigns a high likelihood to sequences of natural-language instructions $\tau_{\text{text}}$ that are plausible ways to reach a final goal $g$ (leveraging textual knowledge embedded in an LLM) which are consistent with visual plans $\tau_{\text{image}}$ starting from image $o$ (leveraging visual dynamics information embedded in a video model), which are further consistent with execution with actions $\tau_{\text{action}}$ (leveraging action information in a large action model). Sampling from this distribution then corresponds to finding sequences $\tau_{\text{text}}, \tau_{\text{image}}, \tau_{\text{action}}$ that are mutually consistent with all constraints, and thus constitute successful hierarchical plans to accomplish the task. 
We provide an illustration of this composition in \fig{fig:hier_composition} with efficacy of this approach demonstrated in~\citep{ajay2023compositional}.

\myparagraph{Controllable Image Synthesis.} Composition can also allows us to construct a generative model $q(x \mid D)$ to generate images $x$ from a detailed scene description $D$  consisting of text and bounding-box descriptions $\{\text{text}_i, \text{bbox}_i \}_{i=1:N}$. This compositional distribution is
\begin{equation*}
    q(x|D) \propto \prod_{i \in \{1, \ldots, N \}} p(x_{\text{bbox}_i} \mid\text{text}_i),
\end{equation*}
where each distribution is defined over bounding boxes in an image.
In \fig{fig:tapestry}, we illustrate the efficacy of this approach for constructing complex images. This approach enables the synthesis of image tapestries~\citep{du2023reduce} and  collages~\citep{zhang2023diffcollage}.

\input{figText/video_composition}

\myparagraph{Style Adaptation of Video Models.} Finally, composition can be used to construct a generative model $q(\tau)$ that synthesizes video in new styles. Given a pretrained video model $p_{\text{pretrained}}(\tau \mid \text{text})$ and a small video model of a particular style  $p_{\text{adapt}}(\tau\mid\text{text})$, we can sample videos $\tau$ from the compositional distribution
\begin{equation*}
    p_{\text{pretrained}}(\tau\mid\text{text}) p_{\text{adapt}}(\tau\mid\text{text})
\end{equation*}
to generate new videos in different specified styles. The efficacy of using composition to adapt the style of a video model is illustrated in ~\citep{yang2023probabilistic}.

%% file: figText/planning_composition.tex
\begin{figure}[t!]
\begin{flushright}
    \raisebox{.35\height}{\rotatebox{90}{\textbf{U-Maze}}}\hspace*{\fill}
    \includegraphics[width=0.95\columnwidth]{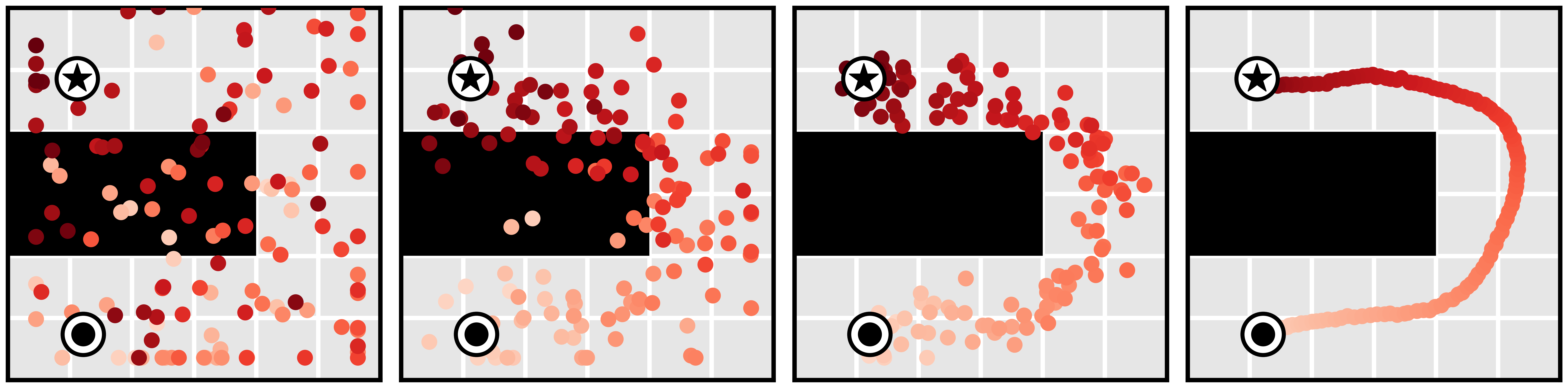} \\
    \raisebox{.3\height}{\rotatebox{90}{\textbf{Medium}}}\hspace*{\fill}
    \includegraphics[width=0.95\columnwidth]{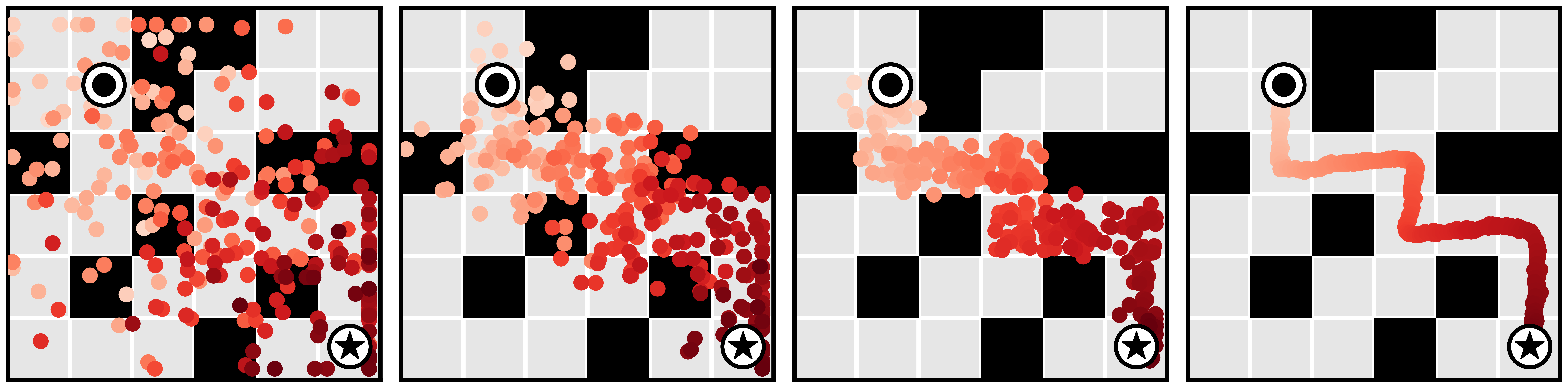} \\
    \raisebox{.6\height}{\rotatebox{90}{\textbf{Large}}}\hspace*{\fill}
    \includegraphics[width=0.95\columnwidth,trim={.03cm 0 .03cm 0},clip]{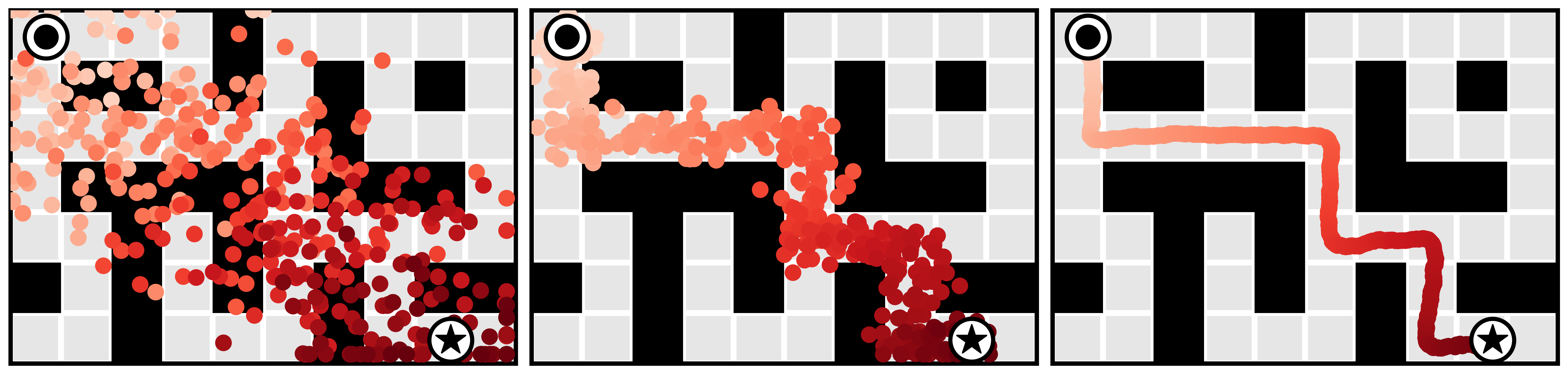} \\
\end{flushright}
\vspace{-10pt}
\caption{
    \textbf{Planning through Probability Composition.} By composing a probability density trained on modeling dynamics in an environment $p_{traj}(\tau)$ with a probability density $p_{goal}(\tau, g)$ which specifies a specific goal state, we can sample plans from specified start 
    \protect{\raisebox{-.05cm}{\includegraphics[height=.35cm]{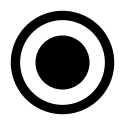}}}
    to a goal
    \protect{\raisebox{-.05cm}{\includegraphics[height=.35cm]{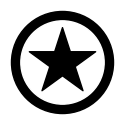}}}
    condition. Figure from ~\cite{janner2022planning}, where the horizontal axis illustrates progression of sampling.
}
\label{fig:compose_plan}
\vspace{-5pt}
\end{figure}

%% file: figText/csp_example.tex
\begin{figure}[t]
    \centering
    \includegraphics[width=\linewidth]{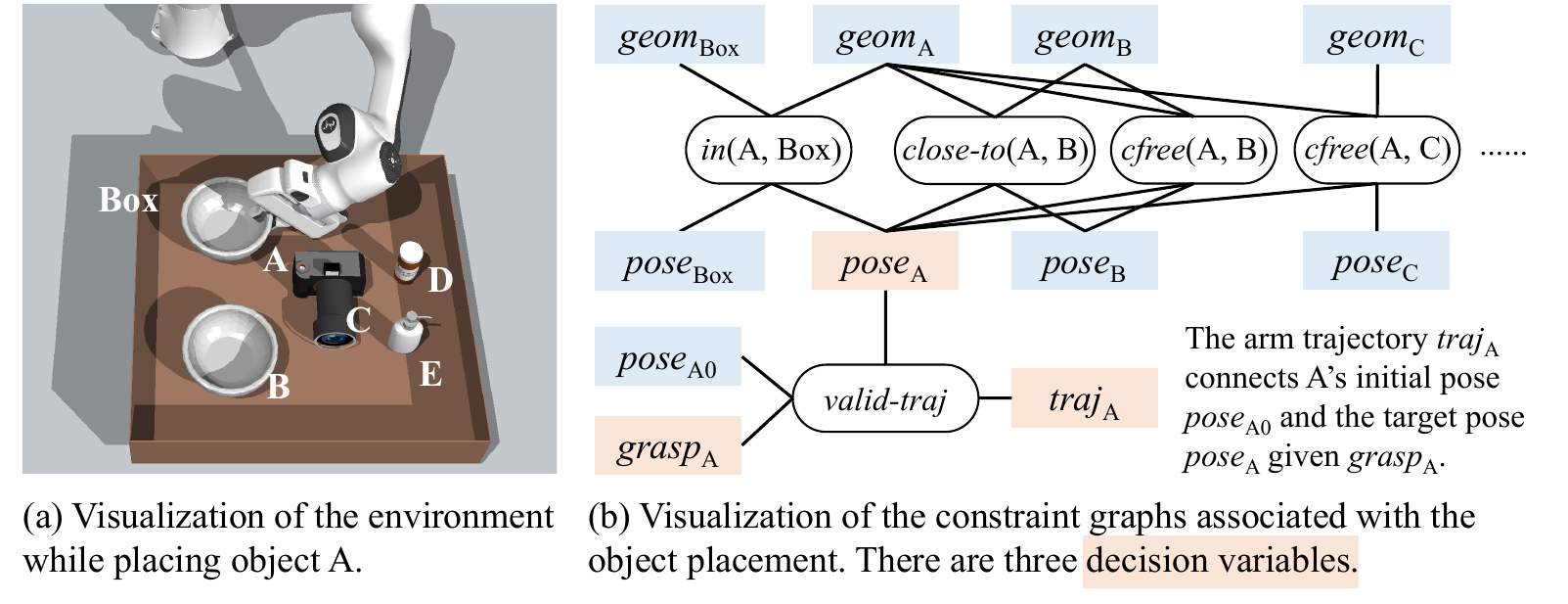}
    \vspace{-20pt}
    \caption{\textbf{Manipulation through Constraint Composition.} New object manipulation problems can be converted into a graph of constraints between variables. Each constraint can be represented as a low-dimensional factor of the joint distribution, with sampling from the composition of distributions corresponding to solving the arrangement problem. Figure adapted from \citep{yang2023compositional}.}
    \label{fig:csp-example}
\vspace{-15pt}
\end{figure}

%% file: figText/hier_composition.tex
\begin{figure}[t]
    \centering
    \includegraphics[width=\linewidth]{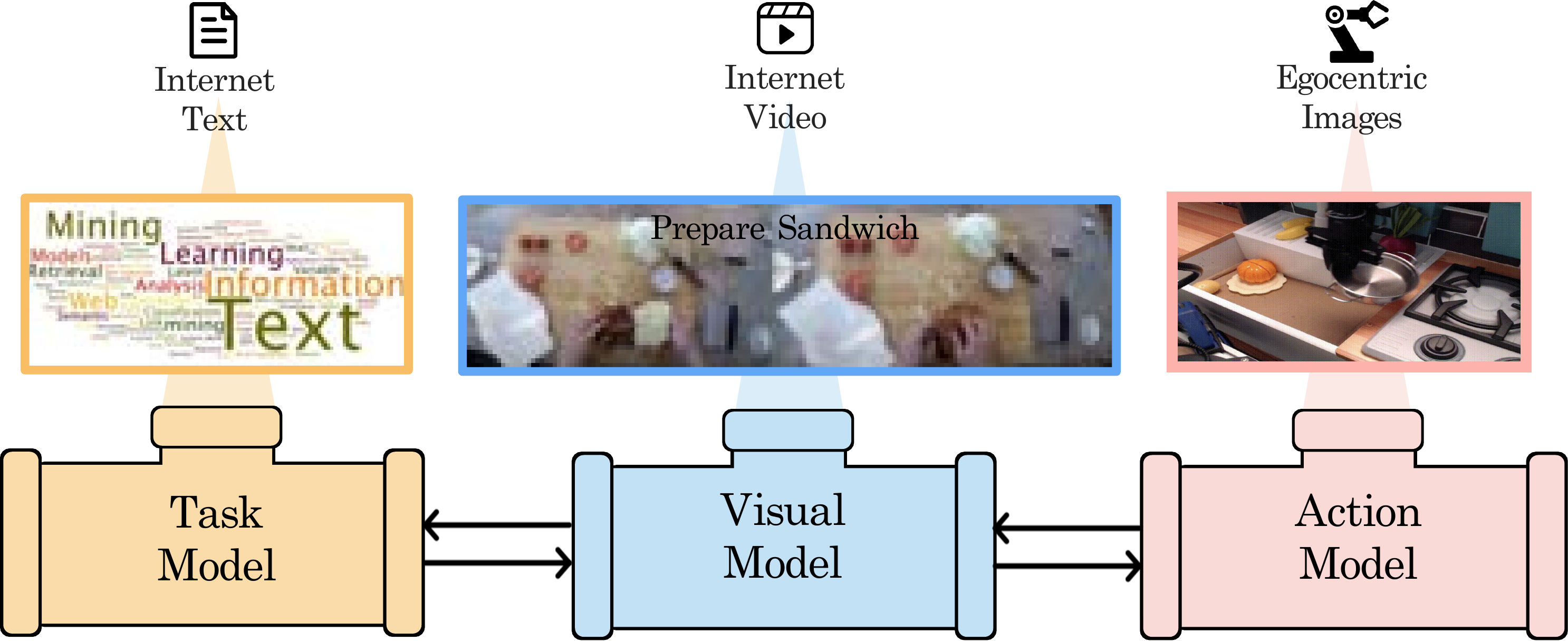}
    \vspace{-20pt}
    \caption{\textbf{Hierarchical Planning through  Composition.}  By composing a set of foundation models trained on Internet data (language, videos, action), we can zero-shot construct a hierarchical planning system. Figure adapted from \citep{ajay2023compositional}.}
    \label{fig:hier_composition}
\end{figure}

%% file: figText/image_tapestry.tex
\begin{figure}[t]
    \centering
    \includegraphics[width=\linewidth]{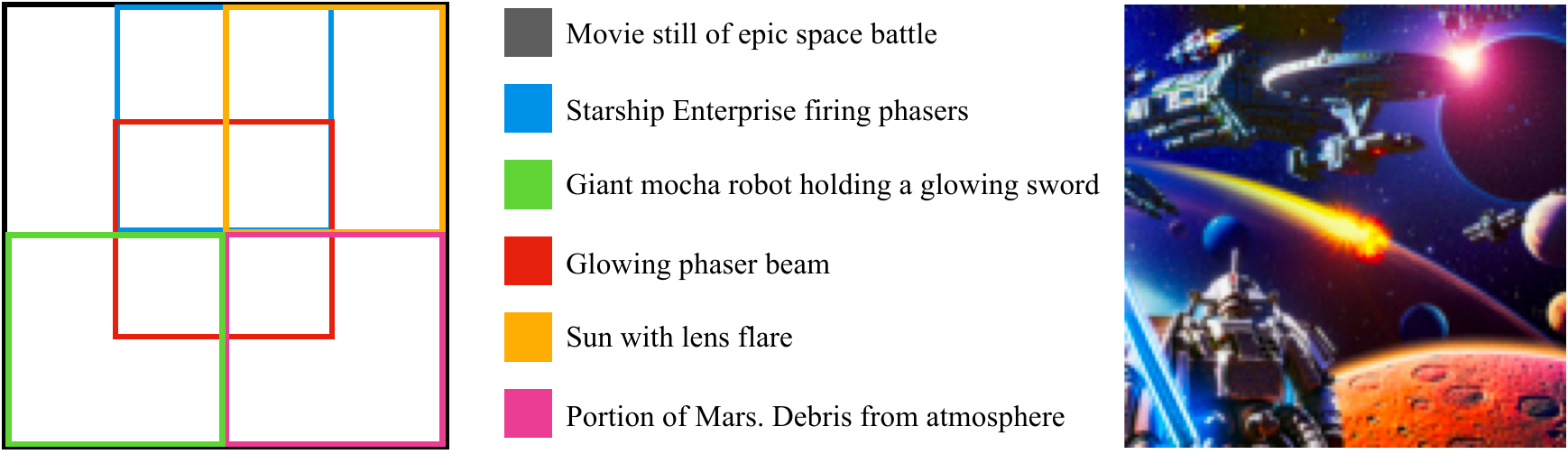}
    \vspace{-20pt}
    \caption{\textbf{Image Tapestries through  Composition.}  By composing a set of probability distributions defined over different spatial regions in an image, we can construct detailed image tapestries. Figure adapted from \citep{du2023reduce}.}
    \label{fig:tapestry}
\vspace{-15pt}
\end{figure}

%% file: figText/video_composition.tex
\begin{figure}[t]
    \centering
    \includegraphics[width=\linewidth]{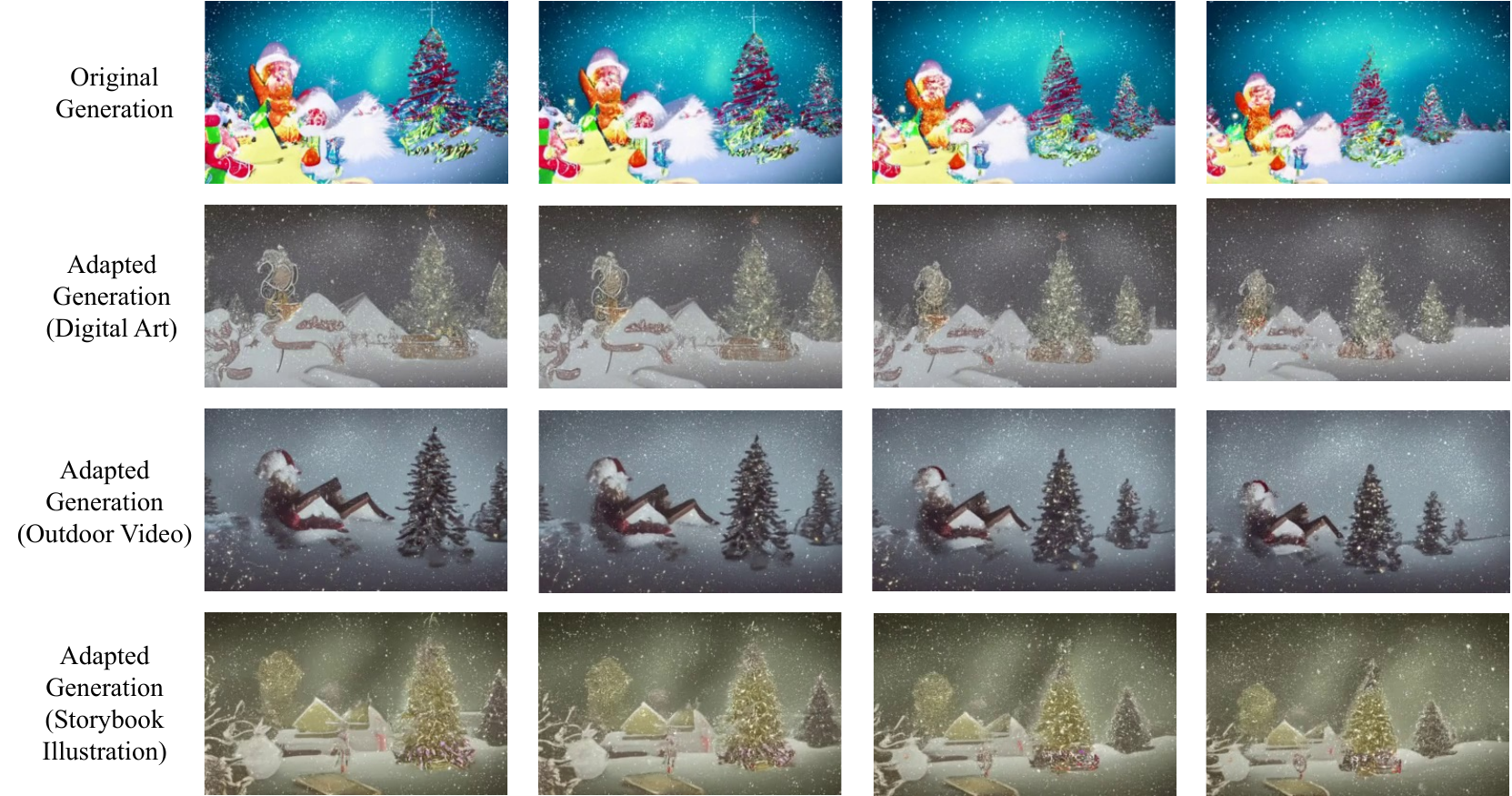}
    \vspace{-20pt}
    \caption{\textbf{Video Stylization through  Composition.}  By composing one video model with a model specifying style,  we can stylize video generations. Figure adapted from \citep{yang2023probabilistic}.}
    \label{fig:video_adapt}
\vspace{-15pt}
\end{figure}

%% file: sections/4_discovering_structure.tex
\section{Generative Modeling with Learned Compositional Structure}

A limitation of compositional generative modeling discussed in the earlier sections is that it requires a priori knowledge about the independence structure of the distribution we wish to model. However, these compositional components can also be discovered jointly while learning a probability distribution by formulating maximum likelihood estimation as maximizing the likelihood of the factorized distribution
\begin{equation*}
    p_\theta(x) \propto \prod_i p_\theta^i(x).
\end{equation*}
Similar to the previous two sections, the discovery of the learned components $p_\theta^i(x)$ enables more data-efficient learning of the generative model as well as the ability to generate samples from new task distributions. Here, we illustrate three examples of how different factors can be discovered in an unsupervised manner.

\input{figText/unsup_composition}

\myparagraph{Discovering Factors from an Input Image.} Given an input image $x$ of a scene, we can parameterize a probability distribution over the pixel values of the image as a product of the compositional generative models
\begin{equation*}
    p_\theta(x) \propto \prod_i p_\theta(x \mid \text{Enc}_i(x)),
\end{equation*}
where $\text{Enc}(\cdot)$ is a learned neural encoder with low-dimensional latent output to encourage each component to capture distinct regions of an image. By training models to autoencode images with this likelihood expression, each component distribution $p_\theta(x\mid \text{Enc}_i(x))$ finds interpretable decomposition of images corresponding to individual objects in a scene as well global factors of variation in the scene such as lighting~\cite{du2021comet,su2024compositional}. In \fig{fig:composition_unsup}, we illustrate how these discovered components, $p_\theta(x\mid z_1)$ and  $p_\theta(x\mid z_2)$ from a model trained on cubes and spheres, $p_\phi(x\mid z_3)$ and  $p_\phi(x\mid z_4)$ from a separate model trained on trucks and boots can be composed together to form the distribution
\begin{equation*}
    p_\theta(x \mid z_1) p_\theta(x \mid z_2) p_\phi(x \mid z_3) p_\phi(x \mid z_4),
\end{equation*}
to construct hybrid scenes with objects from both datasets. 

\input{figText/unsup_relation_composition}

\myparagraph{Discovering Relational Potentials.} Given a trajectory $\tau$ of $N$ particles, we can similarly parameterize a probability distribution over the reconstruction of the particle system as a product of components defined over each pairwise interaction between particles
\begin{equation*}
    p_\theta(\tau) \propto  \prod_{i, j  \forall j \neq i} p_\theta(\tau \mid\text{Enc}_{ij}(\tau)),
\end{equation*}
where $\text{Enc}_{ij}(\tau)$ corresponds to latent encoding interactions between particle $i$ and $j$. In  \fig{fig:unsup_relation}, we illustrate how these discovered relational potentials on one particle system can be composed with relational potentials discovered on a separate set of forces to simulate those forces on the particle system.

\input{figText/unsup_classes_composition}

\myparagraph{Discovering Object Classes From Image Distributions.} Given a distribution of images $p(x)$ representing images drawn from different classes in Imagenet, we can model the likelihood of the distribution as a composition
\begin{equation*}
    p_\theta(x) \propto  p_{\phi}(w\mid x) \prod_i  p_\theta^i(x)^{w_i},
\end{equation*}
where $w_i$ refers to the weighting coefficient for each component.
In \fig{fig:composition_unsup_pretrained}, we illustrate that the discovered components in this setting  represent each of the original Imagenet classes in the input distribution of images. We further illustrate how these discovered components to be composed together to generate images with multiple classes of objects.

%% file: figText/unsup_composition.tex
\begin{figure}[t]
    \centering
    \includegraphics[width=\linewidth]{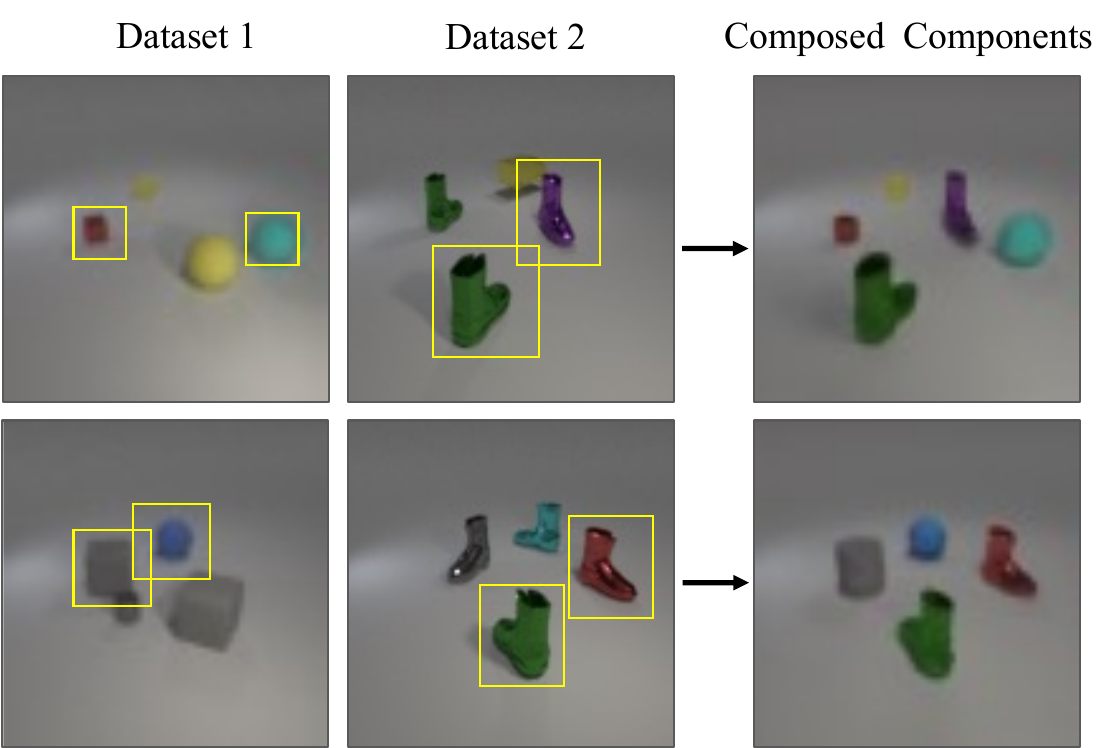}
    \vspace{-15pt}
    \caption{\textbf{Composition of Discovered Objects.}  Probabilistic components corresponding to individual objects in a scene are discovered unsupervised in two datasets using two separate models. Discovered components (illustrated with yellow boxes) can be multiplied together to form new scenes with a hybrid composition of objects. Figure adapted from \citep{su2024compositional}.}
    \label{fig:composition_unsup}
\vspace{-20pt}
\end{figure}

%% file: figText/unsup_relation_composition.tex
\begin{figure}[t]
\centering
  \includegraphics[ trim={30 320 165 0}, width=0.85\textwidth]{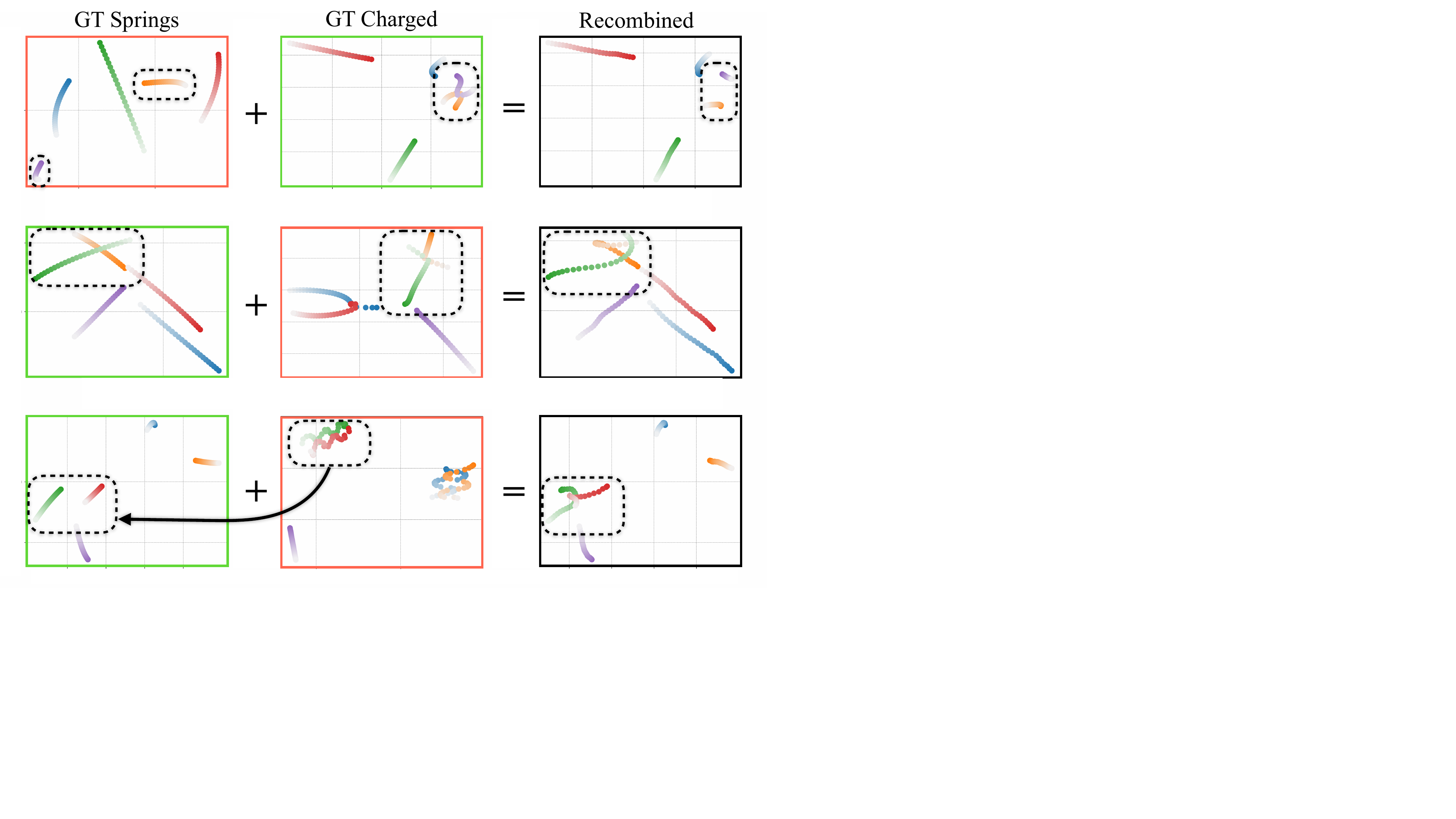}
  \vspace{-20pt}
  \caption{\small \textbf{Composition of Discovered Relation Potentials} In a particle dataset, particles exhibit potentials corresponding to invisible springs between particles (Col. 1) or charges between particles (Col. 2). By swapping discovered probabilistic components between each pair of objects between particle systems, we can recombine trajectories framed in green but with a pair of edge potentials from trajectories formed in red in Col. 3. Figure adapted from ~\citep{comas2023inferring}}
\label{fig:unsup_relation}
\end{figure}

%% file: figText/unsup_classes_composition.tex
\begin{figure}[t]
    \centering
    \includegraphics[width=\linewidth]{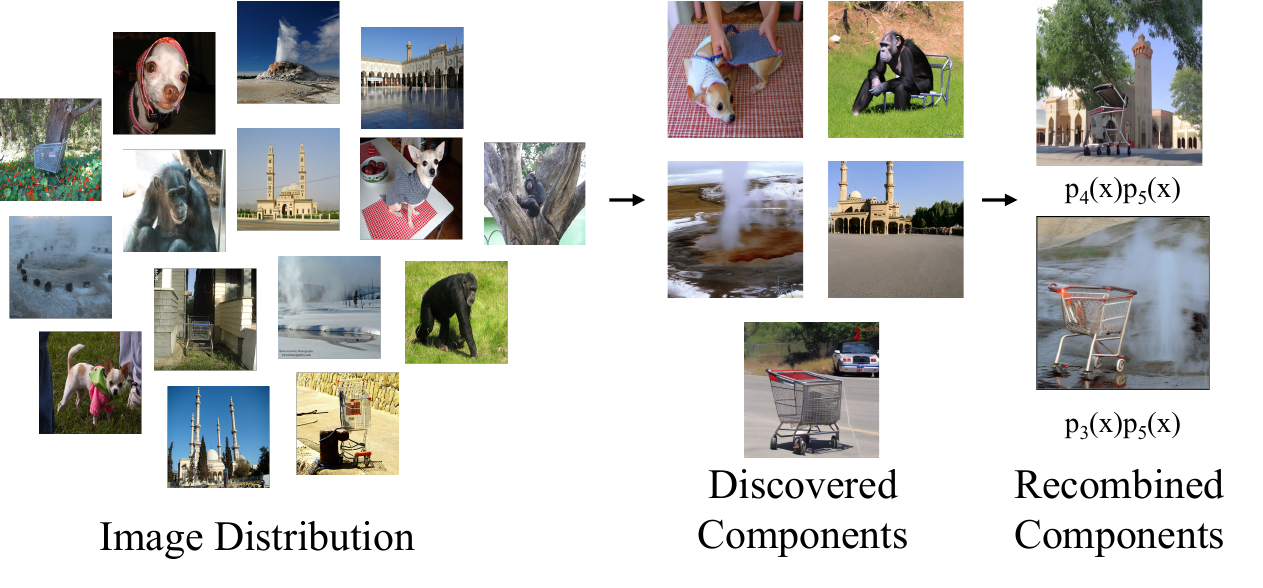}
    \vspace{-25pt}
    \caption{\textbf{Discovering Image Classes.}  Given a distribution of images drawn from 5 image classes in ImageNet,  discovered components correspond to each image class. Components can further be composed together to form new images. Figure adapted from \citep{liu2023unsupervised}.}
    \label{fig:composition_unsup_pretrained}
\vspace{-15pt}
\end{figure}

%% file: sections/5_implementation.tex
\section{Implementing Compositional Generation}

In this section, we discuss some challenges with implementing compositional sampling with common generative model parameterizations and discuss a generative model parameterization that enables effective compositional generation. We then present some practical implementations of compositional sampling in both continuous and discrete domains.

\subsection{Challenges With Sampling from Compositional Distributions}

Given two probability densities $p_1(x)$ and $p_2(x)$, it is often difficult to directly sample from the product density $p_1(x)p_2(x)$. Existing generative models typically represent probability distributions in a factorized manner to enable efficient learning and sampling, such as at the token level in autoregressive models~\citep{van2016pixel} or across various noise levels in diffusion models~\citep{sohl2015deep}. %
However, depending on the form of the factorization, the models may not be straightforward to compose.

For instance, consider two learned autoregressive factorizations $p_1(x_i|x_{0:i-1})$ and $p_2(x_i|x_{0:i-1})$ over sequences $x_{0:T}$. The autogressive factorization of the product distribution $p_{\text{product}}(x) \propto p_1(x) p_2(x)$ corresponds to
\begin{align*}
    p_{\text{product}}(x_i|x_{0:i-1}) = \sum_{x_{i+1:T}} & p_1(x_{i+1:T}|x_{0:i}) p_1(x_i|x_{0:i-1}) \\
     & p_2(x_{i+1:T}|x_{0:i}) p_2(x_i|x_{0:i-1}),
    \end{align*}
where we need to marginalize over all possible future values of $x_{i+1:T}$. Since this marginalization is different dependent on the value of $x_i$, $p_{\text{product}}(x_i|x_{0:i-1})$ is not equivalent to $p_1(x_i|x_{0:i-1}) p_2(x_i|x_{0:i-1})$ and therefore autoregressive factorizations are not directly compositional. Similarly, two learned score functions from diffusion models are not directly composable as they do not correspond to the noisy gradient of the product distribution~\citep{du2023reduce}.

While it is often difficult to combine generative models, representing the probability density explicitly enables us to combine models by manipulating the density. One such approach is to represent probability density as an Energy-Based Model, $p_i(x) \propto e^{-E_i(x)}$~\citep{hinton2002training, du2019implicit}. Under this factorization by definition, we can construct the product density corresponding to 
\begin{equation}
\label{eqn:product}
e^{-(E_1(x)+E_2(x))} \propto e^{-E_1(x)} e^{-E_2(x)},
\end{equation}
corresponding to a new EBM $E_1(x) + E_2(x)$. It is important to observe that EBMs generally represent probability densities  in an unnormalized manner, and the product of two normalized probability densities $p_1(x)$ and $p_2(x)$ will be an unnormalized probability density as well (where the normalization constant is intractable to compute as it requires marginalization over the sample space). Additional operations between probability densities such as mixtures and inversions of distributions can also be expressed as combinations of energy functions~\citep{du2020compositional}. 

To generate samples from any EBM distribution, it is necessary to run Markov Chain Monte Carlo (MCMC) to iteratively refine a starting sample to one that is high likelihood (low energy) under the EBM. We present practical MCMC algorithms for sampling from composed distributions in continuous spaces in \sect{sect:comp_continuous} and discrete spaces in \sect{sect:comp_discrete} with EBMs. Recently, new methods for  implementing compositional sampling using separately trained classifiers to efficiently specify each conditioned factor have been developed~\citep{garipov2023compositional}, which we encourage the reader to also read. 

\subsection{Effective Compositional Sampling on Continuous Distributions}
\label{sect:comp_continuous}

Given a composed distribution represented as EBM $E(x)$ defined over inputs $x \in \mathbb{R}^D$, directly finding a low energy sample through MCMC becomes increasingly inefficient as the data dimension $D$ rises. To more effectively find low-energy samples in EBMs in high-dimensional continuous spaces, we can use the gradient of the energy function to help guide sampling. In ~\citet{du2019implicit},  Langevin dynamics is used to implement efficient sampling, where a sample can be repeatedly optimized using the expression
\begin{equation*}
x_t =  x_{t-1} - \lambda \nabla_x E(x) + \epsilon, \quad \epsilon \sim \mathcal{N}(0, \sigma),
\end{equation*}
where $x_0$ is initialized from uniform noise. By converting different operations such as products, mixtures, and inversions of probability distributions into composite energy functions, the above sampling procedure allows us to effectively compositionally sample from composed distributions~\citep{du2020compositional}.

There has been a substantial body of recent work on improving learning in EBMs~\citep{du2019implicit, nijkamp2019anatomy, grathwohl2019your,du2020improved, grathwohl2021oops} but EBMs still lag behind other generative approaches in efficiency and scalability of training. By leveraging the close connection of diffusion models with EBMs in ~\citep{song2019generative} we can also directly implement the compositional operations with EBMs with diffusion models ~\citep{du2023reduce}, which we briefly describe below.

Given a diffusion model representing a distribution $p(x)$, we can interpret the $T$ learned denoising functions $\epsilon(x, t)$ of the diffusion model as representing $T$ separate EBM distributions, $e^{-E(x, t)}$, where  $\nabla_x{E(x, t)} = \epsilon(x, t)$. This sequence of EBM distributions transition from $e^{-E(x, T)}$ representing the Gaussian distribution $\mathcal{N}(0, 1)$ to $e^{-E(x, 0)}$ representing the target distribution $p_i(x)$. We can draw samples from this sequence of EBMs using annealed importance sampling~\citep{du2023reduce}, where we initialize a sample from Gaussian noise and sequentially run several steps of MCMC on each EBM distribution, starting at $e^{-E(x, T)}$ and ending at $e^{-E(x, 0)}$.

This EBM interpretation of diffusion models allows them to be composed using  operations such as \eqn{eqn:product} by applying the operation to each intermediate EBM corresponding to the component diffusion distributions, for instance $e^{-(E_1(x, k) + E_2(x, k))}$. We can then use an annealed importance sampling procedure on this sequence of composite EBMs. Note that this annealed importance procedure is necessary for accurate compositional sampling -- using the reverse diffusion process directly on this composed score does not sample from the composed distribution~\citep{du2023reduce}.

A variety of different MCMC samplers such as ULA, MALA, U-HMC, and HMC can be used as intermediate MCMC samplers for this sequence of EBM distributions. One easy-to-implement MCMC transition kernel that is easy to understand is the standard diffusion reverse sampling kernel at a fixed noise level. We illustrate in Appendix~\ref{app:ula_reverse} that this is equivalent to running a ULA MCMC sampling step.  This allows compositional sampling in diffusion models to be easily implemented by simply constructing the score function corresponding to the composite distribution we wish to sample from and then using the standard diffusion sampling procedure, but with the diffusion reverse step applied multiple times at each noise level.

\subsection{Effective Compositional Sampling on Discrete Distributions}
\label{sect:comp_discrete}

Given an EBM representing a composed distribution $E(x)$ on a high dimensional discrete landscape, we can use Gibbs sampling to sample from the resultant distribution, where we repeatedly resample values of individual dimensions of $x$ using the marginal energy function $E(x_i \mid x_{-i})$. However, this process is increasingly inefficient as the underlying dimensionality of the data increases. 

The use of a gradient of the energy function $E(x)$ to accelerate sampling in the discrete landscape is difficult, as the gradient operation is not well defined in discrete space (though there are also promising discrete analogs of gradient samplers ~\citep{grathwohl2021oops}). However, we can leverage our learned generative distributions to accelerate sampling, by using one generative model as a proposal distribution and the remaining energy functions to implement a Metropolis-Hastings step ~\citep{li2022composing, verkuil2022language}. 

As an example, to sample from an energy function $E(x) = E_1(x) + E_2(x)$, given an initial MCMC sample $x_t$, we can draw a new sample $x_{t+1}$ by sampling from the learned distribution $e^{-E_1(x)}$, and accept the new sample $x_{t+1}$ with a  Metropolis acceptance rate
\begin{equation*}
    a(x_{t+1}) = \text{clip}(e^{E_2(x_t) - E_2(x_{t+1})}, 0, 1).
\end{equation*}
This procedure allows us to leverage $e^{-E_1(x)}$ to guide sampling from $e^{-E(x)}$ .

%% file: sections/6_discussion.tex
\vspace{-8pt}
\section{Discussion and Future Directions}

\looseness=-1
Most recent research on building generative models has focused on increasing the computational scale and data on which models are trained. We have presented an orthogonal direction to constructing complex generative systems, by building systems {\it compositionally}, combining simpler generative models to form more complex ones. We have illustrated how this can be more data and computation-efficient to learn, enable flexible reprogramming, and how such components can be discovered from raw data.

\input{figText/decentralized_decision_making}
Such compositional systems have additional benefits in terms of both buildability and interpretability. As individual models are responsible for independent subsets of data, each model can be built separately and modularly by different institutions. Simultaneously, at execution time, it is significantly easier to understand and monitor the execution of each simpler constituent model than a single large monolithic model.

In addition, such compositional systems can be more environmentally friendly and easier to deploy than large monolithic models. Since individual models are substantially smaller, they can run efficiently using small amounts of computation. In addition, it is more straightforward to deploy separate models across separate computational machines.

In the setting of constructing an artificially intelligent agent, such a compositional architecture may look like a decentralized decision-making system in \fig{fig:decentralize_decision}. In this system, separate generative models are responsible for processing each modality an agent receives and other models responsible for decision-making. Sampling composed generative distribution of models corresponds to message passing between models, inducing cross-communication between models similar to a set of daemons communicating with each other~\citep{selfridge1988pandemonium}. Individual generative models in this architecture can be substituted with existing models such as LLMs for proposing plausible plans for actions and text-to-video presenting future world states.

Finally, while we have provided a few promising results on applications of compositional generative modeling, there are many limitations to address in future work. First, the current work on compositional modeling assumes a fixed prespecified structure through which models are composed, limiting generalization to new distributions. To flexibly apply compositional models across new tasks, it would be important to construct systems that can instead automatically discover the correct compositional structure between models as well as the appropriate per-model weighting.

Second, current work on discovering compositional structure assumes that data is naturally factorized into an independent product of components. In many real-world settings, gathered data will often exhibit spurious correlations that violate such independence assumptions, causing existing algorithms to fail to discover the correct structure. Exploring more robust approaches to discovering compositional structure such as through prior knowledge or active intervention in the environment are rich directions for future work. 

Lastly, while our focus in this position paper has been on combining separately trained generative models, it would be interesting to theoretically characterize compositional generalization in such systems as well as alternative approaches to improve such generalization. Past theoretical work has characterized compositional generalization in additive models ~\citep{wiedemer2024compositional, lachapelle2024additive}, and it would be interesting to extend such analysis to compositional generative modeling. Furthermore, it would be interesting to explore adding explicit compositional structure to individual models to improve compositional generalization~\citep{misino2022vael,sehgal2023neurosymbolic}.

%% file: figText/decentralized_decision_making.tex
\begin{figure}[t]
\includegraphics[width=\linewidth]{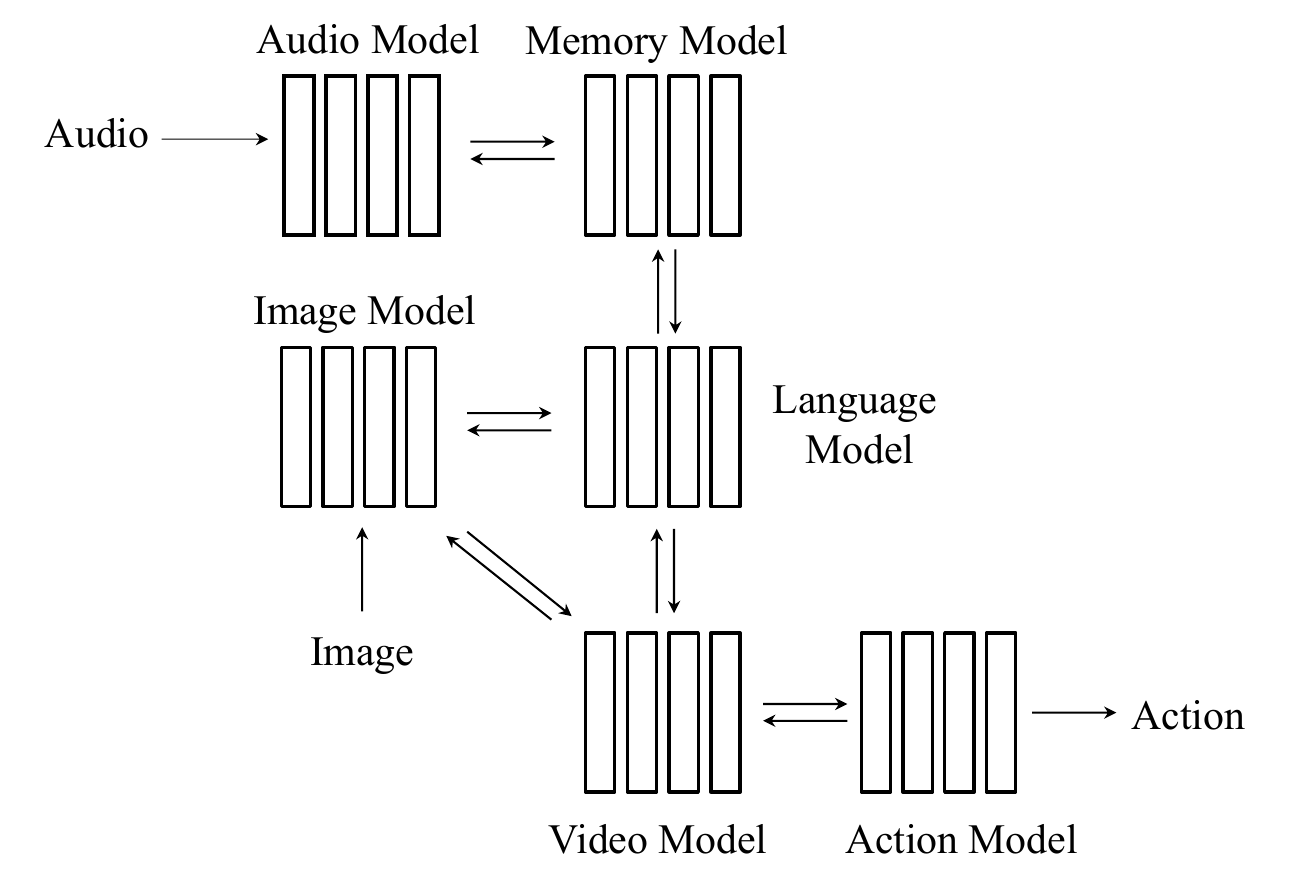}
\vspace{-25pt}
\caption{\small \textbf{Decentralized Decision Making.} By composing generative models operating over various modalities we can construct decentralized architectures for intelligent agents. Communication between models is induced by inference over the joint distribution.}
\label{fig:decentralize_decision}
\vspace{-20pt}
\end{figure}

%% file: sections/7_acknowledgements.tex
\section*{Acknowledgements}
We acknowledge support from NSF grant 2214177; from AFOSR grant FA9550-22-1-0249; from ONR MURI grant N00014-22-1-2740; and from ARO grant W911NF-23-1-0034. Yilun Du is supported by a NSF Graduate Fellowship.

\section*{Impact Statement}
In this paper, we argue that generative models should be built compositionally, from simpler individual parts and illustrate how this enables from data-efficient generative modeling. As generative models become increasingly deployed into production, we believe that such an approach can significantly broaden the impact of such models, enabling them to be deployed in a variety of domains with limited data.

%% file: sections/appendix.tex
\clearpage
\renewcommand{\thesection}{A.\arabic{section}}
\renewcommand{\thefigure}{A\arabic{figure}}
\renewcommand{\thetable}{A\arabic{table}}

\setcounter{section}{0}
\setcounter{figure}{0}
\setcounter{table}{0}

\renewcommand{\theequation}{A\arabic{equation}}
\setcounter{equation}{0}

\appendix
\onecolumn
\textbf{\Large{Appendix}}
\vspace{5pt}

\section{Implementing ULA Transitions as Multiple Reverse Diffusion Steps}
\label{app:ula_reverse}

We illustrate how a step of reverse sampling on a diffusion model at a fixed noise level is equivalent to ULA MCMC sampling at the same fixed noise level. We use the $\alpha_t$ and $\beta_t$ formulation from ~\cite{ho2020denoising}. The reverse sampling step on an input $x_t$  at a fixed noise level at timestep $t$ is given by a Gaussian with a mean 
\begin{equation*}
        \mu_\theta(x_t, t) = x_t - \frac{\beta_t}{\sqrt{1 - \bar{\alpha}_t}}\epsilon_\theta(x_t, t).
\end{equation*}
with the variance of $\beta_t$ (using the variance small noise schedule in~\cite{ho2020denoising}). This corresponds to a sampling update,
\begin{equation*}
        x_{t+1} = x_t - \frac{\beta_t}{\sqrt{1 - \bar{\alpha}_t}}\epsilon_\theta(x_t, t) + \beta_t \xi, \quad \xi \sim \mathcal{N}(0, 1).
\end{equation*}

Note that the expression $\frac{\epsilon_\theta(x_t, t)}{\sqrt{1 - \bar{\alpha}_t}}$ corresponds to the score $\nabla_x p_t(x)$, through the denoising score matching objective~\citep{vincent2011connection}, where the EBM $p_t(x)$ corresponds to the data distribution perturbed with $t$ steps of noise. The reverse sampling step can be equivalently written as
\begin{equation}
        \label{eqn:reverse}
        x_{t+1} = x_t - \beta_t \nabla_x p_t(x) + \beta_t \xi, \quad \xi \sim \mathcal{N}(0, 1).
\end{equation}

The ULA sampler draws an MCMC sample from the EBM probability distribution $p_t(x)$ using the expression
\begin{equation}
        \label{eqn:ula}
        x_{t+1} = x_t - \eta \nabla_x p_t(x) + \sqrt{2} \eta \xi, \quad \xi \sim \mathcal{N}(0, 1),
\end{equation}
where $\eta$ is the step size of sampling.

By substituting $\eta=\beta_t$ in the ULA sampler, the sampler becomes
\begin{equation}
        \label{eqn:ula_beta}
        x_{t+1} = x_t - \beta_t \nabla_x p_t(x) + \sqrt{2} \beta_t \xi, \quad \xi \sim \mathcal{N}(0, 1).
\end{equation}

Note the similarity of ULA sampling in Eqn~\ref{eqn:ula_beta} and the reverse sampling procedure in Eqn~\ref{eqn:reverse}, where there is a factor of $\sqrt{2}$ scaling of the added Gaussian noise in the ULA sampling procedures. This means that we can implement the ULA sampling by running the standard reverse process, but by scaling  the noise added in each timestep by a factor of $\sqrt{2}$. Alternatively,  we can directly we can directly use the reverse sampling procedure in Eqn~\ref{eqn:reverse} to run ULA, where this then corresponds to sampling a tempered variant of $p_t(x)$ with temperature $\frac{1}{\sqrt{2}}$ (corresponding to less stochastic samples from the composed probability distribution).